\documentclass[10pt,journal,compsoc]{IEEEtran}
\ifCLASSOPTIONcompsoc
  \usepackage[nocompress]{cite}
\else
  \usepackage{cite}
\fi
\ifCLASSINFOpdf
\else
\fi
\usepackage{algorithm}
\usepackage{algorithmic}
\usepackage{caption}
\usepackage[utf8]{inputenc} 
\usepackage[T1]{fontenc}    
\usepackage{hyperref}       
\usepackage{url}            
\usepackage{booktabs}       
\usepackage{amsfonts}       
\usepackage{nicefrac}       
\usepackage{microtype}      
\usepackage{graphicx}
\usepackage{subfigure}
\usepackage{amsmath}
\usepackage{amsmath}
\usepackage{color}
\usepackage{bm}
\usepackage{multirow}
\usepackage{float}

\newcommand{\vx}{{\bf x}}

\newcommand{\vz}{{\bf z}}

\newcommand{\vg}{{\bf g}}

\newcommand{\tabincell}[2]{\begin{tabular}{@{}#1@{}}#2\end{tabular}}

\begin{document}

\title{Double Backpropagation for Training Autoencoders against Adversarial Attack}

\author{Chengjin~Sun, Sizhe~Chen, and~Xiaolin~Huang,~\IEEEmembership{Senior~Member,~IEEE}
\IEEEcompsocitemizethanks{\IEEEcompsocthanksitem C.~Sun, S.~Chen, and X.~Huang is with Institute of Image Processing and Pattern Recognition, Shanghai Jiao Tong University, also with the MOE Key Laboratory of System Control and Information Processing, 800 Dongchuan Road, Shanghai, 200240, P.R. China. (e-mails:\{sunchengjin, csz729020210, xiaolinhuang\}@sjtu.edu.cn)\protect\\
\IEEEcompsocthanksitem Corresponding author: Xiaolin Huang.
}
\thanks{Manuscript received 2019.}}

\markboth{}
{Sun \MakeLowercase{\textit{et al.}}: Double Backpropagation for Training Autoencoders}

\IEEEtitleabstractindextext{
\begin{abstract}
Deep learning, as widely known, is vulnerable to adversarial samples. This paper focuses on the adversarial attack on autoencoders. Safety of the autoencoders (AEs) is important because they are widely used as a compression scheme for data storage and transmission, however, the current autoencoders are easily attacked, i.e., one can slightly modify an input but has totally different codes. The vulnerability is rooted the sensitivity of the autoencoders and to enhance the robustness, we propose to adopt double backpropagation (DBP) to secure autoencoder such as VAE and DRAW. We restrict the gradient from the reconstruction image to the original one so that the autoencoder is not sensitive to trivial perturbation produced by the adversarial attack. After smoothing the gradient by DBP, we further smooth the label by Gaussian Mixture Model (GMM), aiming for accurate and robust classification. We demonstrate in MNIST, CelebA, SVHN that our method leads to a robust autoencoder resistant to attack and a robust classifier able for image transition and immune to adversarial attack if combined with GMM.
\end{abstract}
}

\maketitle

\IEEEdisplaynontitleabstractindextext

\IEEEpeerreviewmaketitle

\IEEEraisesectionheading{\section{Introduction}\label{sec:introduction}}

\IEEEPARstart{I}{n} the past few years, deep neural networks have been greatly developed and successfully used in a vast of fields, such as pattern recognition, intelligent robots, automatic control, medicine \cite{alom2018history}. Despite the great success, researchers have found the vulnerability of deep neural networks to adversarial attacks, which has been a big concern recently \cite{akhtar2018threat}. Briefly speaking, for a well-trained neural network $f$ that correctly classifies an example $\vx$, adversarial attacks produce a slight and usually invisible perturbation $\Delta \vx$ such that the output for $\vx + \Delta \vx$ is different from that of $\vx$, which is incorrect. Mathematically, the above adversarial attack on the classifier $f$ can be described as the following problem:
\begin{eqnarray}\label{adv-classifier}
\mathrm{find}  && \Delta\vx \\ \nonumber
\mathrm{s.t.} && f(\vx)\neq f(\vx+\Delta\vx)\\
&& \|\Delta\vx \|\leq \epsilon. \nonumber
\end{eqnarray}
There have been many discussions about adversarial attack since its proposal by Szegedy et al. \cite{szegedy2013intriguing}, which generates adversarial examples with the box-constrained L-BFGS method. The basic idea of (\ref{adv-classifier}) is to find a small magnitude of perturbation to change the output. The intuitive idea is to search on the direction of gradient, which is the basic idea of FGSM \cite{szegedy2013intriguing} and BIM (iterative version of FGSM, \cite{madry2017towards}). Further improvement comes from embedding geometric information, e.g., C\&W \cite{carlini2017towards}, \cite{jia2019comdefend}, Projected Gradient Descent (PGD, \cite{madry2017towards}). The latter is currently regarded as the strongest attack. Many defense methods are also focusing on the difference between the original and the adversarial example on image \cite{raghunathan2018certified,li2018defense}, feature \cite{cisse2017parseval,liao2018defense,prakash2018deflecting,guo2018sparse,yan2018deep}, or network structure \cite{goodfellow2014explaining,szegedy2013intriguing,tramer2017ensemble}.

Adversarial attack and its defense are revolving around a small $\Delta\vx$ and a big resulting difference between $f(\vx + \Delta\vx)$ and $f(\vx)$. In other words, the existence of adversarial examples is rooted in large gradients of the output $f$ with respect to the input $\vx$. Since one usually does not care about $df/d\vx$ when training neural networks, the gradient could become unreasonably large and thus draw forth adversarial attacks. Given this, Drucker and LeCun proposed to restrict $df/d\vx$ by minimizing its magnitude via backpropagation, which is hence named as "double backpropagation" (DBP, \cite{drucker1991double}). This idea has been recently revisited by \cite{ross2018improving} for enhancing the robustness to adversarial attacks. Although the basic idea of pursuing small gradients for robustness sounds promising, DBP has not achieved the expectation for classification tasks. This is because in classification tasks, the label $y$ is discontinues, so forcing the neural network to have small $df/d\vx$ may lead to $f$'s inability to approach $y$.

In this paper, we are going to use DBP in training autoencoders. For an autoencoder, the encoder part first encodes the input into a lower dimensional code $\vz=e(\vx)$. Then the decoder reconstructs the input from the latent code $\vx^{\prime}=d(\vz)$. Unlike the discontinuity of labels, the latent space of an encoder should be continuous with respective to the input, i.e., if the inputs are similar, their latent representations, and reconstruction results should be similar.

However, to the best of our knowledge, the existing autoencoders do not control the magnitude of the $d\vx^{\prime}/d\vx$, which leads autoencoders to be easily attacked. The attack on an autoencoder in output space can be described as the following,

\begin{eqnarray}\label{adv-autoencoder}
\mathrm{find}  && \Delta\vx \\ \nonumber
\mathrm{s.t.}&& \|d(e(\vx)) - d(e(\vx+\Delta\vx))\| \geq D_d \nonumber \\
&& \|\Delta\vx \|\leq \epsilon. \nonumber
\end{eqnarray}
where $e(\cdot)$ and $d(\cdot)$ are the encoder and decoder to be attacked.

\begin{figure}[htbp]
	\centering
	\includegraphics[width=\hsize]{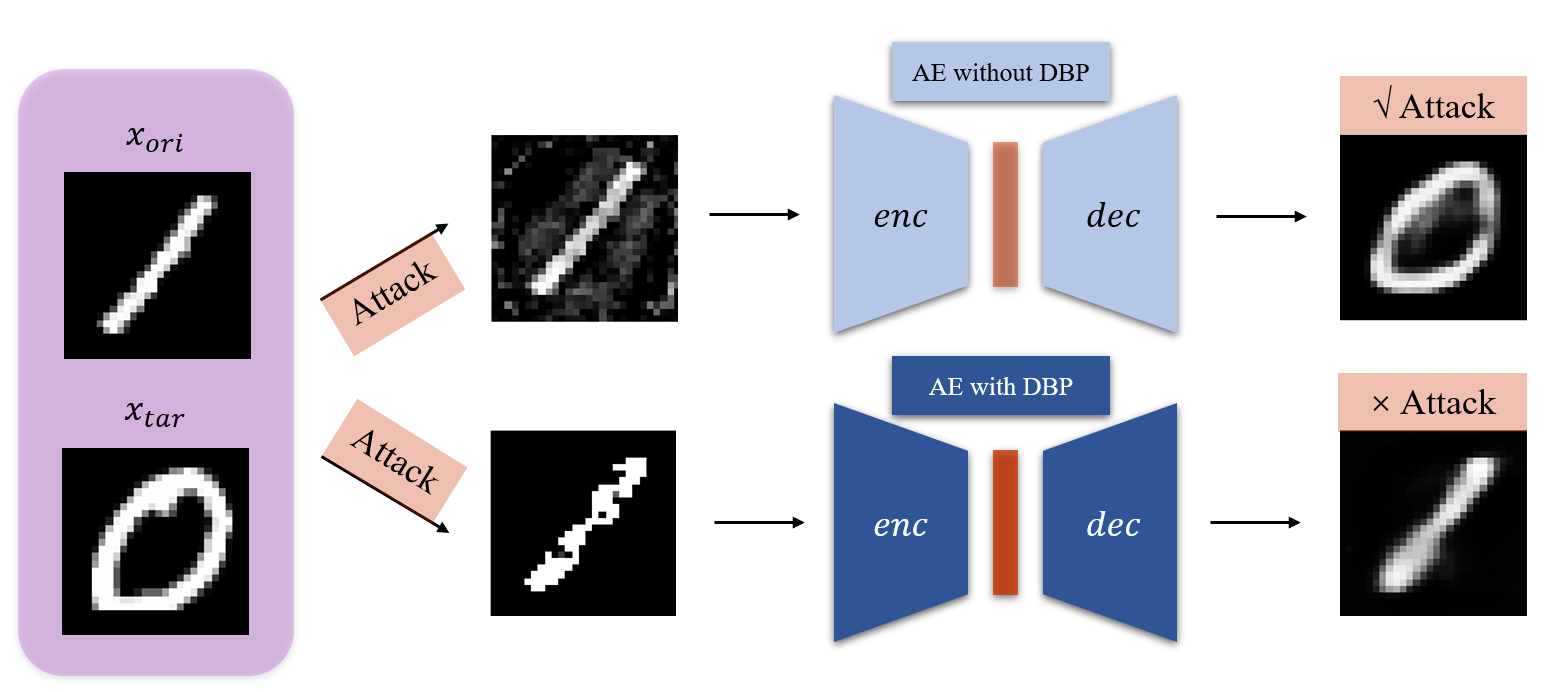}
	\caption{An example of adversarial attack on DRAW.
		The left column shows attack target (from "1" to "0"). The adversarial example of AE (DRAW) and the reconstruction results are shown in the top line. The adversarial example looks like the original "1", however, after encoding and decoding, the reconstruction result, shown in the top right, becomes to another digit "0". The bottom line shows the performance of DBP. For the same model and the same attack, the reconstruction result of the adversarial example still resembles digit "1".}
	\label{f1}
\end{figure}

Notice that $\vx$ is a clean example so the distance between $\vx$ and its reconstructed output $\vx^{\prime}$ is within a reconstruction error bound $D$, i.e., $\|d(e(\vx)) - \vx\| \leq D$. Then according to the requirement of (\ref{adv-autoencoder}), we could find that the reconstructed output of adversarial example is different from the input. Instead, it could resemble another target image or could be totally meaningless, the former of which is targeted attack and the latter is non-target. An example of target attack is shown in top of Fig. \ref{f1}, where the left gives the original image and target image and the top-right plots the reconstruction result of an adversarial example. The attacked autoencoder is Deep Recurrent Attentive Writer (DRAW) \cite{gregor2015draw} and the attack is the one proposed by \cite{gondim2018adversarial}. With this attack, one can slightly disturb an original input to obtain an adversarial input, that is similar to the origin but its reconstruction is totally different.

The autoencoders' safety is of great importance, because they are widely used as compression schemes for data storage and transmission. Given that the attack results from the large gradients of autoencoders, we propose to use DBP to control the gradient and improve their robustness to adversarial attacks.

Therefore we propose to train an autoencoder by minimizing the reconstruction loss, the KL-divergence, as the existing training process, and the magnitude of the gradients. By training with DBP, an autoencoder with small $d\vx^{\prime}/d\vx$ could be obtained, which prevents a slight perturbation on $\vx$ to lead significant difference on the latent space and the reconstruction result. For the previous example shown in Fig. \ref{f1}, when we training DRAW with DBP, the attack fails, i.e., the reconstruction result resembles the input.


The above describes the basic idea training autoencoders by DBP. In the following sections, we specifically train DBP autoencoders in the framework of Variational Autoencoder (VAE) \cite{kingma2013auto,rezende2014stochastic} and DRAW. The experiments are on MNIST \cite{lecun1998gradient} and CelebA \cite{liu2015deep} dataset and AUDDC is used to measure their robustness. Comparison results show that autoencoder trained with DBP is more robust against adversarial attacks, in both visual effect and quantitative measurements.

Via training with DBP, we can obtain an autoencoder with controllable gradients $d\vz/d\vx$ and $d\vx^{\prime}/d\vx$. According to that, we further design a distribution estimator based on GMM, which attempts to reconstruct continuous probability $p$ from discrete labels $y$. This procedure also could be regarded as label smoothing \cite{muller2019does,hinton2015distilling,papernot2016distillation,szegedy2016rethinking,geng2013facial}. In the proposed GMM, $dp/d\vz$ is kept small and its cooperation with an encoder trained by DBP leads to a classifier with reasonable gradients from the output to the input, which is hence robust to attacks. A prominent example is that going alone the obtained classifier, one could transfer the images rather than generate adversarial examples which is an example for application of double backpropagation.

The remaining parts of the paper are organized as follows. In Section II, we give a brief review of autoencoders and describe their vulnerability under adversarial attacks. Section III describes the design of double backpropagation for autoencoders. Section IV discusses the cooperation of autoencoder with DBP and GMM. Experiments that validate the robustness of the proposed DBP are given in Section V. Section VI concludes the paper briefly.

\section{Autoencoder and Its Attack}
A typical autoencoder consists of two parts, encoder $e$ which compresses the input into a low dimensional latent representation, and decoder $d$ which reconstructs the input from the latent space. Autoencoders and their variations have been widely applied \cite{gondim2018adversarial}. There have been many different encoding and decoding architectures, e.g., sparse \cite{ng2011sparse}, denoising \cite{vincent2010stacked}, variational \cite{kingma2013auto, rezende2014stochastic}, Wasserstein \cite{tolstikhin2017wasserstein}, symmetric \cite{pu2017adversarial}. In this paper, we focus on VAE and DRAW. The former is a classical autoencoder framework and the latter is a newly designed one. We implement double backpropagation on these two models as a representative study for their variants.

VAE is a powerful deep generative model, which is effective in generating data in many complicated distributions. 
VAE aims to maximize $p_\theta(x)$, the distribution to generate $x$ from latent variables $z$ as follows,
\begin{equation}
p_\theta(x)=\int p_\theta(x,z)dz=\int p_\theta(x|z)p(z)dz.
\end{equation}
In VAE, an arbitrary distribution could be used in the latent space. But in practice, the standard Gaussian distribution is often employed \cite{kingma2013auto}, i.e., $q(z) = \mathcal N(0,1)$. In VAE training process, we optimize the following lower bound for $p_{\theta}(x)$.
\begin{equation}
\begin{aligned}
{\rm log} \ p_{\theta}(x) \geq E_{q_\phi(z|x)}[{\rm log} \ p_{\theta}(x|z)]-D_{KL}(q_{\phi}(z|x)||p(z))
\end{aligned}
\end{equation}
$q_\phi(z|x)$ is the encoder part which encodes $x$ into $z$, and $p_{\theta}(x|z)$ represents decoder which reconstructs $x'$ from $z$. $\theta$ and $\phi$ are corresponding parameters.

DRAW is a newly proposed structure with a sequential variational auto-encoding framework that allows for the iterative construction of complex images. DRAW uses LSTMs with an attention mechanism and a metaphor of "painting" the image in a canvas step by step \cite{gondim2018adversarial}. The training procedure could be summarized as the following for $t =1,2,...,T$
\begin{equation}
\begin{aligned}
\hat{x}_t&=x-\sigma(c_{t-1})\\
r_t&={\rm read}(x_t,\hat{x}_t,h_{t-1}^{\rm dec})\\
h_{t}^{\rm enc}&={\rm RNN}^{\rm enc}(h_{t-1}^{\rm enc},[r_t,h_{t-1}^{\rm dec}])\\
z_t&\sim Q(Z_t|h_{t}^{\rm enc})\\
h_{t}^{\rm dec}&={\rm RNN}^{\rm dec}(h_{t-1}^{\rm dec},z_t)\\
c_t&=c_{t-1}+{\rm write}(h_{t}^{\rm dec}),
\end{aligned}
\end{equation}
where $\hat{x}_t$ denotes the error image, $\sigma$ is the logistic sigmoid function, $c_{t}$ represents the canvas matrix, $h_{t}^{\rm enc}$ denotes the output of ${\rm RNN}^{\rm enc}$, and $h_{t}^{\rm dec}$ denotes the output of ${\rm RNN}^{\rm dec}$. $h_{t}^{\rm dec}$ is used to parameterize the distribution $Q(Z_t|h_{t}^{\rm enc})$, with which the parameters of the Gaussian distribution in the latent can be calculated as follows,
\begin{equation}
\begin{aligned}
\mu_t&=W(h_{t}^{\rm enc})\\
\sigma_t&=exp(W(h_{t}^{\rm enc}))
\end{aligned}
\end{equation}
The total loss $L$ of DRAW can be represented as:
\begin{equation}
\begin{aligned}
L&=-{\rm log} \ D(x|c_T)\\
&+\frac{1}{2}\left(\sum_{t=1}^{T}\nolimits\mu_t^2+\sigma_t^2-{\rm log} \ \sigma_t^2\right)-T/2\\
\end{aligned}
\end{equation}
Here, $c_T$ is the final canvas matrix. The negative log probability of $x$ under $D$ is $-{\rm log} \ D(x|c_T)$, which represents reconstruction loss and the other loss is for $z$.

Since the proposal of adversarial attack \cite{goodfellow2014explaining}, its design, defense, and analysis have attracted much attention recently. Except for a new type of adversarial attack \cite{tang2019adversarial}, the majority of the existing adversarial attacks aim at the over-sensitive part of a neural network such that slight distortions on the input lead to significant changes on the output.

As discussed previously, the gradients of autoencoders have not been considered during its training neither, thus the existence of adversarial examples is not surprising. Here, we introduce the targeted attack on autoencoders designed by \cite{gondim2018adversarial}. It aims to change the input from $\vx_0$ within a small range such that the reconstruction output is similar to a target $\vx_{\mathrm{tar}}$. Mathematically,
\begin{eqnarray}\label{adv-encod-1}
\min_{\Delta \vx} & & \|\Delta \vx\| + \lambda \| d(e(\vx_0 + \Delta \vx))- \vx_{\mathrm{tar}} \|.
\end{eqnarray}
where $\lambda$ is a trade-off parameter between the similarity to the target and the restriction on the change. In some point of view, the code is more important to an autoencoder and then the distance is measured in the latent space, i.e.,
\begin{eqnarray}\label{adv-encod-2}
\min_{\Delta \vx} & & \|\Delta \vx\| + \lambda \| e(\vx_0 + \Delta \vx)- e(\vx_{\mathrm{tar}}) \|.
\end{eqnarray}
Notice that the norm here could be any distance measurement, e.g., pixel-wise $\ell_2$ distance for the image space and KL-divergence for the latent space.

The robustness of autoencoders against adversarial attacks could be measured by the area under the distortion–distortion curve (AUDDC) defined in \cite{gondim2018adversarial}. 

\section{Autoencoder with double back-propagation}
For an autoencoder, two similar inputs should share similar codes ideally, which actually implies that the magnitude of $d\vx^{\prime}/d\vx = d\vx^\prime/d\vz \cdot d\vz/d\vx $ should be small. Therefore, we use double backpropagation scheme to train autoencoders. Concisely, we minimize $d\vx^{\prime}/d\vx$ together with the reconstruction loss and the latent space loss, i.e.,
\begin{equation}\label{DBP-auto}
\begin{aligned}
\arg\min_{e, d} L_{\mathrm{rec}}(d(e(\vx)), \vx) + \lambda_1 L_{\mathrm{lat}}(e(\vx)) +\\ \lambda_2 ||\nabla_\vx L_{\mathrm{rec}}(d(e(\vx)), \vx)||_2^2,
\end{aligned}
\end{equation}
where $L_{\mathrm{rec}}$ stands for the reconstruction loss, $L_{\mathrm{lat}}$ for the latent loss in the latent space. $\lambda_1$ and $\lambda_2$ are hyper parameters specifying the penalty strength. In (\ref{DBP-auto}), the autoencoder could be smooth and imperceptible perturbation of the input cannot significantly change the latent codes and the reconstruction output, which greatly improves the robustness of autoencoders against adversarial attacks.

When training autoencoders, different architectures should adopt different learning rate and $\lambda_2$. Large learning rate and $\lambda_2$ lead to a divergent reconstruction result. In controlling proportions of different item, the total loss focuses on the loss of the reconstruction in the first half of the time and loss of DBP in the other. 

The DBP scheme for training an autoencoder is previously defined by (\ref{DBP-auto}). For a specific model, the structures of $e$, $d$, the loss for reconstruction, and the latent space loss are different. First, we consider VAE(DBP), which can be represented as:
\begin{equation}\label{DBP-VAE}
\begin{aligned}
\mathop{\arg\min}_{\phi,\theta}  &-E_{q_\phi(z|x)}[{\rm log} \ p_{\theta}(x|z)]+\lambda_1 D_{KL}(q_{\phi}(z|x)||p(z)) \\
&+\lambda_2 \left\|\nabla_x -E_{q_\phi(z|x)}[{\rm log} \ p_{\theta}(x|z)] \right\|_2^2.\\
\end{aligned}
\end{equation}

In practice, we simultaneously train the encoder and the decoder to minimize the (\ref{DBP-VAE}). The encoder $e(\cdot)$ maps the input $x$ to a Gaussian distribution with its mean and variance to be $\mu(x,\phi)$ and $\sigma(x,\phi)$. Then a latent variable z is sampled from such Gaussian
distribution by the reparameterization trick \cite{kingma2013auto}, which is used for recovery. $\phi$ and $\theta$ represent the parameter of encoder and decoder respectively. $\lambda_1$ and $\lambda_2$ are the hyper parameters that balance the reconstruction performance and the magnitude of the gradient. $\Gamma$ denotes gradient descent method which is used for updating training parameters $\phi$ and $\theta$. This optimization algorithm is described in Algorithm \ref{alg1},
\begin{algorithm}
	\caption{Training VAE with DBP}
	\label{alg1}
	\begin{algorithmic}
		\REQUIRE $learning \ rate, \lambda_1, \lambda_2$
		\STATE $epoch\leftarrow 0$
		\WHILE{$epoch< 500$}
		\STATE $x\leftarrow {\rm GetMiniTtrainingBatch}()$
		\STATE $z \sim \mathcal N(\mu(x,\phi),\sigma(x,\phi))$
		\STATE $(\phi,\theta) \leftarrow (\phi,\theta)+ \Gamma(\frac{\partial}{\partial\phi}(L_0+\lambda_1\cdot L_1
		+\lambda_2\cdot L_2),  \frac{\partial }{\partial\theta}(L_0$ \STATE $+\lambda_1\cdot L_1+\lambda_2\cdot L_2) )$
		\ENDWHILE
	\end{algorithmic}
\end{algorithm}
where $L_0=-E_{q_\phi(z|x)}[{\rm log} \ p_{\theta}(x|z)]$,
$L_1=D_{KL}(q_{\phi}(z|x)||p(z))$,
$L_2=||\nabla_x -E_{q_\phi(z|x)}[{\rm log} \ p_{\theta}(x|z)]||_2^2$.

For DRAW with double back-propagation, the optimization objective function becomes
\begin{equation}
\begin{aligned}
\mathop{\arg\min}_{\phi,\theta} &=-{\rm log} \ D(x|c_T)+\lambda_1\left(\frac{1}{2}\left(\sum_{t=1}^{T}\mu_t^2+\sigma_t^2-{\rm log} \ \sigma_t^2\right)\right.\\
&\left. -T/2\right) +\lambda_2 \|\nabla_x -{\rm log} \ D(x|c_T) \|_2^2,
\end{aligned}
\end{equation}
where $\lambda_1=1$ and
$$ \lambda_2=\left\{
\begin{array}{rcl}
0       &      & {epoch < TotalEpoch/2}\\
1     &      & {epoch > TotalEpoch/2}\\
\end{array} \right. $$

At each time step $t$, the encoder maps the input from the image $x$, the previous hidden vector $h^{\rm dec}_{t-1}$ and $read$ operation. Then a latent variable $z_t$ is sampled from the Gaussian distribution $\mathcal N(\mu(x,\phi_t),\sigma(x,\phi_t)$. The decoder receives $z_t$ and previous hidden vector $h^{\rm dec}_{t-1}$ to get the output $h^{\rm dec}_{t}$ which is added (via a $write$ operation) to $canvas$ matrix $c_t$. After $T$ steps, the final canvas matrix $c_T$ is used to reconstruct the image. So the reconstruction loss is $-{\rm log} \ D(x|c_T)$ and the DBP loss is $||\nabla_x -{\rm log} \ D(x|c_T) ||_2^2$. The training parameters are $\phi$ and $\theta$, which are accumulation of $\phi_t$ and $\theta_t$ at each time step.
This optimization algorithm to train DRAW with DBP is described in Algorithm \ref{alg2},
\begin{algorithm}
	\caption{Training DRAW with DBP}
	\label{alg2}
	\begin{algorithmic}
		\REQUIRE $learning \ rate, \lambda_1, \lambda_2,T,c_0,
		h_{0}^{\rm enc},h_{0}^{\rm dec}$
		\STATE $epoch\leftarrow 0$
		\WHILE{$epoch< 500$}
		\STATE $x\leftarrow {\rm GetMiniTtrainingBatch}()$
		\WHILE{$t< T$}
		\STATE $\hat{x}_t=x-\sigma(c_{t-1})$
		\STATE $r_t={\rm read}(x_t,\hat{x}_t,h_{t-1}^{\rm dec})$
		\STATE $h_{t}^{\rm enc}={\rm RNN}^{\rm enc}(h_{t-1}^{\rm enc},[r_t,h_{t-1}^{\rm dec}])$
		\STATE $\mu_t=W(h_{t}^{\rm enc})$
		\STATE $\sigma_t={\rm exp}(W(h_{t}^{\rm enc}))$
		\STATE  $z_t \sim \mathcal N(\mu_t(x,\phi_{t}),\sigma_t(x,\phi_{t}))$
		\STATE $h_{t}^{\rm dec}={\rm RNN}^{\rm dec}(h_{t-1}^{\rm dec},z_t)$
		\STATE $c_t=c_{t-1}+{\rm write}(h_{t}^{\rm dec})$
		\ENDWHILE
		\STATE $\phi=(\phi_{0},\phi_{1},...,\phi_{T}),\theta=(\theta_{0},\theta_{1},...,\theta_{T})$
		\STATE $(\phi,\theta) \leftarrow (\phi,\theta)+$ \STATE $ \Gamma(\frac{\partial}{\partial\phi}(L_0+\lambda_1 \cdot L_1
		+\lambda_2 \cdot L_2), \ \frac{\partial}{\partial\theta}(L_0+\lambda_1 \cdot L_1
		+\lambda_2 \cdot L_2))$
		\ENDWHILE
	\end{algorithmic}
\end{algorithm}
where $L_0=-{\rm log} \ D(x|c_T)$, $L_1=\frac{1}{2}(\sum_{t=1}^{T}\mu_t^2+\sigma_t^2-{\rm log} \ \sigma_t^2)-T/2$, and $L_2=||\nabla_x -{\rm log} \ D(x|c_T) ||_2^2$.

\begin{figure}[htbp]
	\centering
	\includegraphics[width=\hsize]{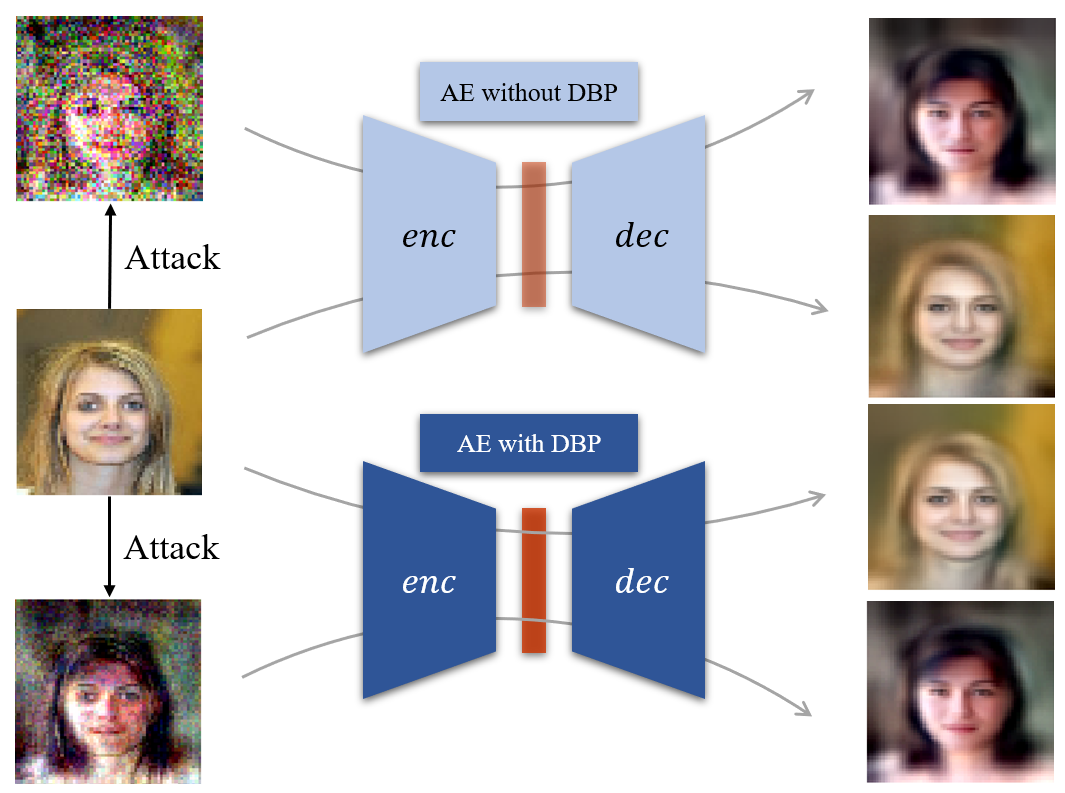}
	\caption{Attack performance on VAE(NO DBP) and VAE(DBP). From the original image in the middle of the left column, one can generate an adversarial example (shown in the top) for which the reconstruction result (shown in the top of the right column), is totally different. If we train VAE with DBP, the reconstruction result for the original image is similar to that for VAE; see the middle two images in the right column, but the attack fails, i.e., the adversarial example (the bottom left image) and its reconstruction result (the bottom right image) are similar.
	}
	\label{f14}
\end{figure}

Adopting DBP in training of autoencoders could certainly decrease the sensitivity. Here, we give an example in Fig.\ref{f14} to illustrate the difference. The top row of it shows examples of attacking VAE without DBP. Though the input is similar to the original example, its output is totally different, i.e., very similar to the target. As explained, this phenomenon results from very large gradient $d\vx^{\prime}/d\vx$ in the original VAE, which is 131.874. With DBP, we successfully control the gradients to 7.356, nearly 1/12 of the original one, while keeping the reconstruction error similar, which is 2.771 for VAE(NO DBP) and 2.796 for VAE(DBP).
The comparative images could be found in the bottom row of Fig.\ref{f14}, where the reconstruction for clean image is quite good and the attack fails. To produce output similar to target, the input sample transforms to something like the target already, which goes against the principle of adversarial attack. 

\section{Smooth Classifier via Encoder with DBP and GMMs}
The idea of "double backpropagation" has been introduced for a long time but its performance on classification is far from satisfactory. The basic reason is that
the labelling process changes continuous probabilities $p$ to discrete labels $y$,
 e.g., $-1$ and $+1$ in binary classification. And it is not reasonable to require neural networks that approach the labels to have small gradients. In this paper, we propose to train autoencoder with double backpropagation which ensures $d\vz/d\vx$ to be small and meaningful. Therefore, to establish a smooth classifier $f$, one should first recover the probability to make the other part $df/d\vz$ to be reasonable, for which the recently proposed idea, label smoothing, \cite{szegedy2016rethinking} may help. For example, soft labels acquired by a distillation mechanism \cite{papernot2016distillation} could improve the robustness of model by re-training \cite{mustafa2019image}. Label smoothing also has remarkable advantages for obtaining good within-class distributions, e.g., \cite{geng2013facial,muller2019does}. The above label smoothing methods only focus on changing the discrete label to distribution to make the $df/d\vx$ more meaningful. However, we can utilize the smooth encoder to compress the input to a low dimensional space to make the label smoothing process simpler.

In this paper, we model the latent space $Z$ by Gaussian Mixture Models (GMM) to learn distribution instead of discrete labels. The idea of establishing GMM in the latent has been proposed in clustering \cite{guo2017deep,alqahtani2018deep} and novel detection \cite{zong2018deep}. With GMMs in a latent space, we could keep $df/d\vz$ small and reasonable. 
Furthermore, $df/d\vx=df/d\vz \cdot d\vz/d\vx$ is smooth and meaningful with smooth encoder, so we obtain a classifier expected to be robust.

Suppose there are $N$ classes in a classification problem. We establish $N$ GMMs in $\vz$ to fit them. For the $i^{th}$ GMMs, the distribution of $\vz$ can be represented as:
\begin{equation}
p_i(\vz|\pi_i,\bm\mu_i,\bm\Sigma_i)=\sum_{j=1}^K\pi_{ij}\mathcal{N}(\vz|\bm\mu_{ij},\bm\Sigma_{ij}
)
\end{equation}
$\pi_i$, $\bm\mu_i$, $\bm\Sigma_i$ are parameters of $i^{th}$ class.
$K$ is the number of components of every GMM. $\pi_{ij}$ denotes $j^{th}$ component's prior probability in $i^{th}$ GMM. Similarly, $\bm\mu_{ij}$ is mean vector and $\bm\Sigma_{ij}$ is covariance matrix. Suppose the dataset of $i^{th}$ class is $X_i=\{\vx_{i1},\vx_{i2},...,\vx_{iM_{i}}\}$. We first train an autoencoder with double backpropagation, and then the encoder compresses $X_i$ to  $Z_i=\{\vz_{i1},\vz_{i2},...,\vz_{iM_{i}}\}$ in the latent. Then Expectation-Maximization (EM) algorithm \cite{moon1996expectation} is adopted to estimate the parameters in every GMM including prior probability $\pi_{ij}$, mean vector $\bm\mu_{ij}$ and covariance matrix $\bm\Sigma_{ij}$ based on $Z_i$.

After learning parameters of GMM, the probability density value belonging to every class can be calculated for any point in latent space. We concatenating and normalize them to get the probability that the point belongs to each class, which we regard as the output of a classifier.

\begin{equation}
\vg(\vz)=[ p_1^{\prime}(\vz),p_2^{\prime}(\vz)…p_N^{\prime}(\vz)],\quad p_i^{\prime}(\vz)\in[0,1],\quad\sum_{i=1}^N p_i^{\prime}=1
\end{equation}
Combining the encoder part and the GMM part, we get the smooth classifier $f(\vx) = \vg(e(\vx))$. So, the transformation between two classes can be implemented by running several iterations of the following equations.
\begin{equation}
\label{Transition}
\vx_{new}=\vx-\eta\cdot sign(\bigtriangledown_{\vx}\parallel f_{target}-f(\vx)\parallel_2)
\end{equation}
where $f_{target}$ is the target output and $\eta$ denotes the step size.

\begin{figure}[htbp]
	\centering
	\includegraphics[width=\hsize]{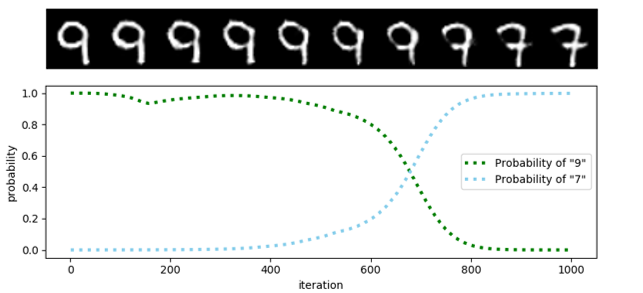}
	\caption{By modeling a GMM over an autoencoder with DBP, one can obtain a smooth classifier such that changing the image in the gradient direction could really transfer the image from "9" and "7", which also means that adversarial attack in gradient direction fails. The bottom plots the smooth change of the probabilities of "9" and "7".
}
	\label{f0}
\end{figure}

As shown in the following \emph{transformation}, new examples are generated based on gradient as same to the adversarial attack. However, instead of obtaining noisy images with incorrect label, we successfully transfer the image, the probability, and the label jointly, from "9" to "7". Figure \ref{f0} illustrates the images and the probabilities.

Transformation and adversarial attacks do not co-exist. The success of transformation implies the robustness of the established GMM. Therefore, the classifier we obtained by GMM estimation and smooth encoder is relatively robust to adversarial attacks. Gaussian mixture variational autoencoder \cite{ghosh2019resisting} is similar to our model in the structure. Nevertheless, this defense method does not maintain a good reconstruction result. Furthermore, it also fails in white box when the hyper parameter that trades off between reconstruction fidelity and latent space prior changes.

\section{Experiment}
As previous discussed, training autoencoder with DBP could restrict the magnitude of the gradient so that improve the performance against adversarial examples. In this section, we validate the robustness of proposed method by securing autoencoder against adversarial attack. First, we train autoencoder with DBP and validate that it improves robustness against adversarial examples greatly comparing the autoencoder without DBP on MNIST and CelebA dataset.
The autoencoders we use are VAE and DRAW, where adopting "double backpropagation" method makes a difference in defending. After that, we train an autoencoder with DBP and build GMMs in latent space to obtain robust classifier on MNIST and SVHN. Then we validate its transferability and robustness against white and black attack.


\subsection{Reconstruction Performance}
The autoencoders we choose to research are VAE and the DRAW without and with its attention mechanism. For autoencoders' structure, we adopt the models in \cite{gondim2018adversarial}. The details about the models and their implementation are given in the supplemental materials.

\begin{figure}[!h]
	\centering
	\includegraphics[width=\hsize]{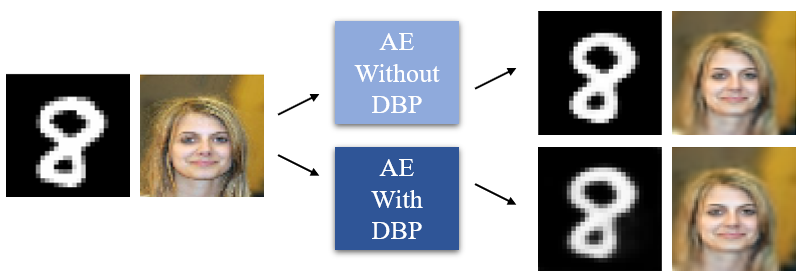}
	\caption{Reconstruction results of DRAW without and with DBP trained on MNIST and CelebA. The left column shows two original images and the reconstruction results are given in the right, indicating that DBP only slightly discards the reconstruction performance.
}
	\label{f2}
\end{figure}

\begin{table*}
	\caption{Reconstruction loss and magnitude of gradient($l_2$ norm) for all autoencoders of different architecture with and without double back-propagation on MNIST and CelebA.}
	\centering
	\begin{tabular}{l|l|l|l|l|l|l|l|l|l}
		\toprule
		&&\multicolumn{2}{c}{VAE} & \multicolumn{2}{c}{DRAW1} & \multicolumn{2}{c}{DRAW16} & \multicolumn{2}{c}{DRAW16(attention)}\\

		&& DBP & NO DBP & DBP & NO DBP & DBP & NO DBP & DBP & NO DBP \\
		\hline
		\multirow{2}{*}{Reconstruction Loss} &MNIST  &0.238 &0.319 &0.259 &0.297 &0.174 &0.237 &0.133 &0.168        \\
		&CelebA &2.771 &2.796 &3.017 &3.068 &2.612 &2.561 &1.875 &1.813  \\

		\midrule
		\multirow{2}{*}{Magnitude of Gradient}&MNIST  &9.535 &0.051 &6.732 &0.041 &10.056 &0.045 &10.657 &0.023       \\
		&CelebA &133.274 &7.541 &579.255 &10.057 &652.772 &2.660 &1061.100 &13.992  \\
		\bottomrule
	\end{tabular}
	\label{reconstruct}
\end{table*}
Table \ref{reconstruct} shows that "double back propagation" restricts the magnitude of gradient $d\vx^{\prime}/d\vx$, most of which reaches nearly 1/100 of the original one. At the same time, the quality of reconstruction does not decrease too much.
In Fig. \ref{f2}, we show reconstruction results of AE(DBP) and AE(NO DBP) on MNIST and CelebA for the qualitative analysis. For each pair, the left image is an example from MNIST and the right one is from CelebA.

\subsection{Robustness of VAE with DBP}
In the previous part, we show that applying DBP in the autoencoder can restrict the amplitude of $d\vx^{\prime}/d\vx$ significantly without decreasing the reconstruction performance. Naturally, we expect that the smoother autoencoder could improve the resistance to adversarial attack. So in this part, we evaluate the robustness of VAE with DBP comparing with VAE without DBP. We achieve targeted attack both on the latent space and output space.

\begin{figure}[!h]
	\centering
	\includegraphics[width=\hsize]{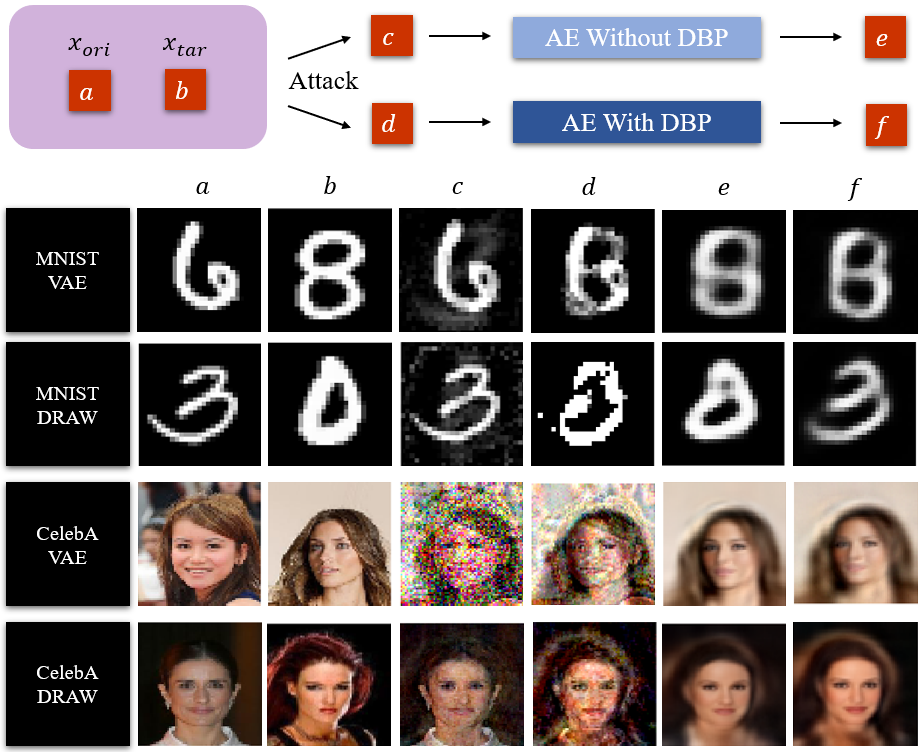}
	\caption{Robustness of autoencoder against attacks. The top shows the attack process: a. the original image; b. the target image; c. adversarial example of autoencoder without DBP; d. adversarial example of autoencoder with DBP; e. reconstruction result when there is no DBP; f. reconstruction result when there is DBP. The left column indicates the specific autoencoder (VAE/DRAW) and the dataset (MNIST/CelebA). With additionally considering DBP, the robustness of autoencoders is visually enhanced such that unless the adversarial example is very similar to the target, the reconstruction result can not approach the target image. 
}
	\label{AE}
\end{figure}

Here AUDDC is used to evaluate the robustness of autoencoders. The details about calculating the AUDDC are provided in supplemental material. Table \ref{AUDDC} shows VAE(NO DBP) and VAE(DBP) performance on adversarial attacks. The AUDDC of VAE(DBP) is much larger than that of VAE(NO DBP), which means applying double back-propagation in VAE successfully improves the robustness against adversarial examples.
Then we visually analyzed the attack results for VAE(DBP) and VAE(NO DBP) on each dataset. Fig. \ref{AE} shows the attack on MNIST and CelebA dataset respectively. For MNIST, when attacking VAE(NO DBP), the adversarial examples are generated by adding imperceptible perturbations on the original image, leading the output to change to something similar to the target. The adversarial examples of VAE(DBP) are always similar to the target, which means the restriction of the gradient $d\vx^{\prime}/d\vx$ improves the resistance of the VAE to attack. For CelebA, when the reconstruction results approach the target, the adversarial examples of VAE(NO DBP) are noisy image but they are similar to target in VAE(DBP).

\subsection{Robustness of DRAW with DBP}
In this part, we evaluate the robustness of DRAW(DBP) and DRAW(NO DBP) on MNIST and CelebA dataset. The architectures of DRAW are different in attention window, time steps, size of latent and the LSTM. We achieve targeted attack both on the latent space and output space. 

Table \ref{AUDDC} illustrates the performance of these autoencoders against adversarial attacks. The AUDDCs of different DRAWs(DBP) are much higher than that of DRAW(NO DBP), which means for autoencoders with more complex architecture such as DRAW, applying double back-propagation can also strengthen the resistance against adversarial examples.
In Fig. \ref{AE}, we show some examples of the attack. Imperceptible additive perturbations on the input image fool the DRAW(NO DBP) to generate something similar to the target. However, adversarial examples can only fool the DRAW(DBP) when they actually look like the target.
\begin{table*}
	\caption{AUDDC for DRAW different structure with and without DBP on MNIST and CelebA.
		Higher values indicate higher resistance to the attacks.}
	\label{AUDDC}
	\centering
	\begin{tabular}{l|l|l|l|l|l|l|l|l|l}
		\toprule
		
		&&\multicolumn{2}{c}{VAE} & \multicolumn{2}{c}{DRAW1} & \multicolumn{2}{c}{DRAW16} & \multicolumn{2}{c}{DRAW16(attention)}\\

		&& DBP & NO DBP & DBP & NO DBP & DBP & NO DBP & DBP & NO DBP \\
		\hline
		\multirow{2}{*}{$A_{latent}$} & MNIST &0.304 & 0.55 &0.264 &0.529&0.409 &{0.614} &0.693 & 0.801  \\

		&CelebA &0.329 &0.429 &0.453 &0.578 &0.858 &0.865 &0.934 &0.946     \\
		
		\hline
		\multirow{2}{*}{$A_{output}$} & MNIST &0.277 &0.447  &0.239 &0.402 &0.143 &0.323 &0.361 &0.605   \\
		&CelebA &0.238 &0.369 &0.342 &0.485 &0.198 &0.319 &0.232 &0.310     \\
		\bottomrule
	\end{tabular}
	\label{t3}
\end{table*}

\subsection{Image Transition based on GMM}
The above experiments validate the robustness of equipping double backpropagation on VAE and DRAW. As previously discussed, training AE(DBP) can control the magnitude of $d\vz/d\vx$. We build GMMs in the latent space to make $df/d\vz$ meaningful, so as for $df/d\vx$. Then we validate its transferability and robustness against while and black attack.
In this part, we consider the dataset MNIST and SVHN.  The networks we choose to generate adversarial examples are a simple MLP, LeNet \cite{lecun1998gradient} and designed convolutional network CvNet. Attack methods are FGSM and PGD. The details about the networks are provided in supplemental material.

For MNIST dataset,   
we experiment with two classifiers $f_1,f_2$ for demonstration of DBP. $f_1$ combines the autoencoder trained without DBP and with GMM. $f_2$ applies DBP on the autoencoder whose $\lambda_1$ and $\lambda_2$ in (\ref{DBP-auto}) are both 1. We then build GMM in the latent space for classification. 
\begin{figure}[htbp]
	\centering
	\includegraphics[width=\hsize]{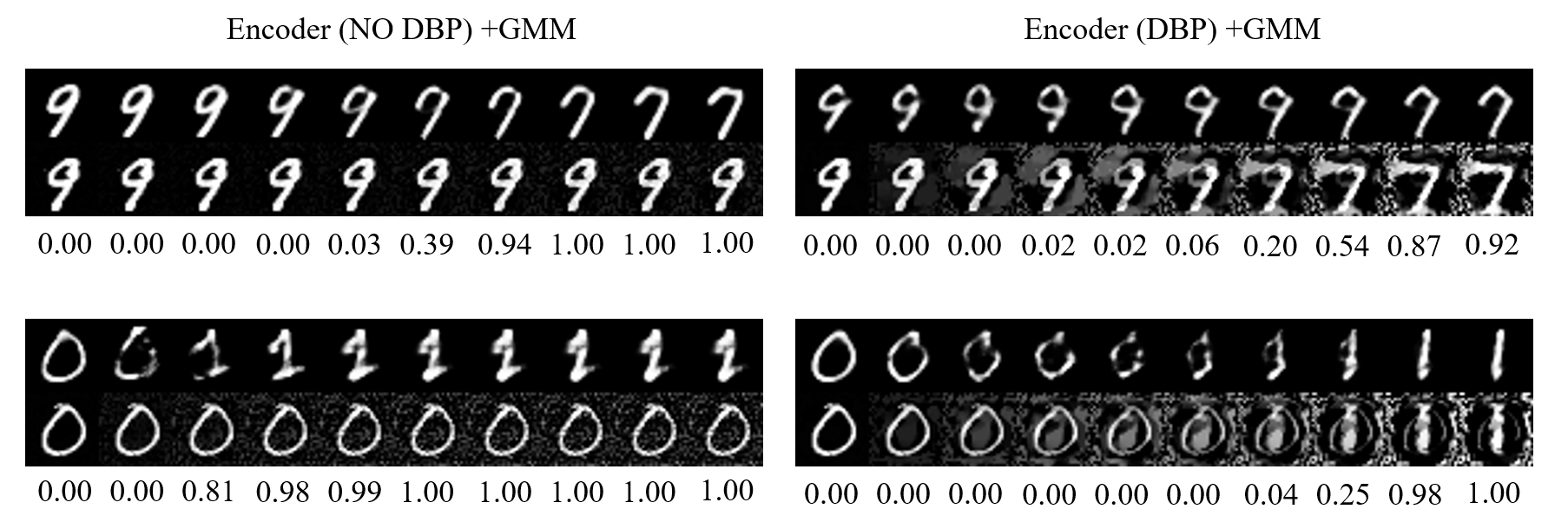}
	\caption{Image transition performance with and without DBP. In each of the four groups, we transfer the images by searching on the gradient direction of the classifier. The first line is the reconstruction result of the autoencoder and the second line is the input. 
}	\label{f7}
\end{figure}
Four examples of transition in the input space are shown in Figure \ref{f7}. During the transition, the confidence increases gradually. Transition in the input space of $f_1$ is shown in the left of Figure \ref{f7}. The confidence increases and output of decoder transforms to the target gradually, but the input just has some noise and does not change to the target. Then we implement transition in the input space of $f_2$ . The results are visualized in the right of Figure \ref{f7}. The input changes greatly and shows some apparent features of the target as the confidence changing for the reason that the GMM is smooth and the magnitude of $d\vz/d\vx$ is restricted.

So transformation in the input space can only be realized when both the GMMs and encoder are smooth. However, the input images are still fuzzy and hard to recognize because the GMM does not reach high accuracy and encoder part is not smooth enough. GMM may provide the wrong gradient direction which leads the input change to fuzzy one.

As previous discussed, transformation and adversarial attacks do not co-exist. To validate this, we analyze the performance of the classifier we design against white and black box examples. For MNIST, MLP and LeNet are used to generate black box examples \cite{madry2017towards} where the accuracy of MLP is 97.84\% and LeNet is 99.37\%. The classifier with encoder and GMMs reaches 95.28\% accuracy rate. For SVHN, we train CvNet to generate specific adversarial examples which reaches 91.14\% accuracy rate. The classifier consisting of encoder and GMM reaches 83.8\% accuracy.
Table \ref{White Attack} illustrates the performance of classifiers against white attack. We find that the classifiers combining smooth encoder and GMM shows robustness against the FGSM and PGD attack while the MLP, LeNet and CvNet can be easily attacked.
\begin{table}[t]
	\caption{Accuracies of the Encoder(DBP)+GMM, MLP, LeNet and CvNet for the white box examples on MNIST and SVHN. 
	}
	\centering
	\begin{tabular}{llllll}
		\toprule
		& \multicolumn{3}{c}{MNIST}&\multicolumn{2}{c}{SVHN} \\
		\midrule
		&\tabincell{c}{Encoder(D\\BP)+GMM}      & MLP      &  LeNet & \tabincell{c}{Encoder(D\\BP)+GMM} &CvNet\\
		\midrule
		FGSM     & 94.83\% & 19.03\%         & 91.48\%& 80.0\%& 58.9\%\\
		PGD     & 93.56\%      & 8.12\%     & 56.64\% & 72.6\% & 1.8\%\\
		\bottomrule
	\end{tabular}
	\label{White Attack}
\end{table}
The classifiers combining smooth encoder and GMM can also defend the adversarial black box attack which is illustrated in Table \ref{Black Attack}.
\begin{table}[H]
	\caption{Accuracies of the Encoder(DBP)+GMM for the black box examples on MNIST and SVHN. 
		The adversarial examples are generated by MLP, LeNet and CvNet.
	}
	\centering
	\begin{tabular}{llll}
		\toprule
		& \multicolumn{2}{c}{MNIST}&SVHN \\
		\midrule
		& MLP     & LeNet      &CvNet   \\
		\midrule
		FGSM     & 93.33\% & 92.32\%         & 82.0\%\\
		PGD     & 92.93\%      & 88.77\%     &73.0\% \\
		\bottomrule
	\end{tabular}
	\label{Black Attack}
\end{table}

\section{Conclusion}
In this paper, we are the first to propose to secure autoencoder such as VAE and DRAW with double backpropagation. It is shown that adversarial attack towards an autoencoder trained with DBP will fail and produce sample similar to the target image. Accordingly, autoencoders with DBP are much more robust without loss of reconstruction reality. Furthermore, we obtain a robust and accurate classifier by modeling the distribution of latent vectors in robust autoencoder by GMM. Our research shows that smoothing the gradient and label contributes to defense of adversarial attack.

\ifCLASSOPTIONcompsoc
  \section*{Acknowledgments}
\else
  \section*{Acknowledgment}
\fi
This work was partially supported by National Key Research Development Project (No. 2018AAA0100702), National Natural Science Foundation of China (No. 61977046), and 1000-Talent Plan (Young Program).
\ifCLASSOPTIONcaptionsoff
  \newpage
\fi

\bibliographystyle{IEEEtran}
\bibliography{Reference}

\vspace{-1cm}
\begin{IEEEbiography}[{\includegraphics[width=1in,height=1.25in,clip,keepaspectratio]{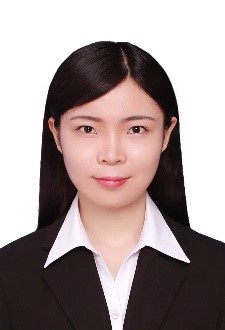}}]{Chengjin Sun}
received her BS degree in Nanjing University, Nanjing, China, in 2018. She is now a master student at the Institute of Image Processing and Pattern Recognition, Shanghai Jiao Tong University, Shanghai, China. Her research interests are adversarial robustness for deep learning.
\end{IEEEbiography}

\begin{IEEEbiography}[{\includegraphics[width=1in,height=1.25in,clip,keepaspectratio]{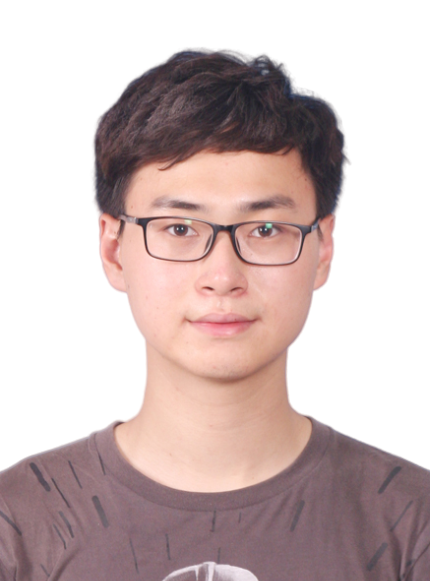}}]{Sizhe Chen}
is a senior student in Department of Automation, Shanghai Jiao Tong University, Shanghai, China. He is now doing research at the Institute of Image Processing and Pattern Recognition, Shanghai Jiao Tong University. His research interests are adversarial attack, attack transferability, and interpretability of DNN.
\end{IEEEbiography}

\begin{IEEEbiography}[{\includegraphics[width=1in,height=1.25in,clip,keepaspectratio]{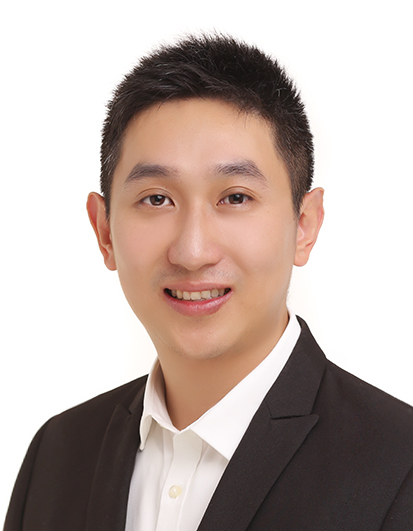}}]{Xiaolin Huang}
(S'10-M'12-SM'18) received the B.S. degree in control science and engineering, and the B.S. degree in applied mathematics from Xi'an Jiaotong University, Xi'an, China in 2006. In 2012, he received the Ph.D. degree in control science and engineering from Tsinghua University, Beijing, China. From 2012 to 2015, he worked as a postdoctoral researcher in ESAT-STADIUS, KU Leuven, Leuven, Belgium. After that he was selected as an Alexander von Humboldt Fellow and working in Pattern Recognition Lab, the Friedrich-Alexander-Universit\"{a}t Erlangen-N\"{u}rnberg, Erlangen, Germany. From 2016, he has been an Associate Professor at Institute of Image Processing and Pattern Recognition, Shanghai Jiao Tong University, Shanghai, China. In 2017, he was awarded by "1000-Talent Plan" (Young Program). His current research areas include machine learning and optimization.
\end{IEEEbiography}

\end{document}